\documentclass[pmlr]{jmlr}% new name PMLR (Proceedings of Machine Learning)

\usepackage{longtable}% for long tables

\usepackage{booktabs}
\usepackage[load-configurations=version-1]{siunitx} % newer version
 \usepackage{color}
 \usepackage{amsmath}
 \usepackage{mathtools}
 
 \let\given\givenbase
 \usepackage{amsfonts}
 \usepackage[T1]{fontenc}
\newcommand{\norm}[1]{\left\lVert#1\right\rVert}

\theorembodyfont{\upshape}
\theoremheaderfont{\scshape}
\theorempostheader{:}
\theoremsep{\newline}

\jmlryear{}
\jmlrworkshop{
}

\title[DRL Survey for Clinical Decision Support]{Deep Reinforcement Learning for Clinical Decision Support: A Brief Survey}

\begin{document}
%\author{Liu Siqi}
\author{\Name{Siqi Liu} \Email{E0272316@u.nus.edu} \\
       \addr NUS Graduate School for Integrative Sciences and Engineering\\
       National University of Singapore\\
       21 Lower Kent Ridge Rd, Singapore 119077
       \AND
       \Name{Kee Yuan Ngiam} \Email{kee\_yuan\_ngiam@nuhs.edu.sg}\\
       \addr Department of General Surgery\\National University Hospital\\
       5 Lower Kent Ridge Rd, Singapore 119074
       \AND
       \Name{Mengling Feng} \Email{ephfm@nus.edu.sg}\\
       \addr Saw Swee Hock School of Public Health\\National University of Singapore\\
       12 Science Drive 2, Singapore 117549}
\maketitle
\begin{abstract}
  Owe to the recent advancements in Artificial Intelligence especially deep learning, 
  many data-driven decision support systems have been implemented to facilitate medical doctors in delivering personalized care. 
  We focus on the deep reinforcement learning (DRL) models in this paper.
 DRL models have demonstrated human-level or even superior performance in the tasks of computer vision and game playings, such as Go and Atari game. 
 However, the adoption of deep reinforcement learning techniques in clinical decision optimization is still rare.  
 We here present the first survey that summarizes reinforcement learning algorithms with Deep Neural Networks (DNN) on clinical decision support.
 We also discuss some case studies, where different DRL algorithms were applied to address various clinical challenges. We further compare and contrast the advantages and limitations of various DRL algorithms and present a preliminary guide on how to choose the appropriate DRL algorithm for particular clinical applications.
\end{abstract}

\section{Introduction}
%Deep learning is a promising new technique to relieve human from labor-intensive feature engineering processes. 
%
The effective combination of deep learning (deep neural networks) and reinforcement learning technique, named Deep Reinforcement Learning (DRL), is initially invented for intelligent game playing\citep{mnih2013playing,silver2016mastering} and has later emerged as an effective method to solve complicated control problems with large-scale, high-dimensional state and action spaces\citep{mnih2015human}. The great applications of deep reinforcement learning to these domains have relied on the knowledge of the underlying processes (e.g., the rules of the game).\\

In the healthcare domain, the clinical process is very dynamic. The 'rules' for making clinical decisions are usually unclear\citep{marik2015demise}. For instance, let us consider the following clinical decision-making questions: will the patient benefit from the organ transplant; under what condition the transplant will become a better option; what are the best medications and dosages to prescribe after the transplantation. The decisions from those questions should dedicate to the individual patient's condition. In order to find the optimal decisions to those questions, conducting randomized clinical trials (RCTs) are usually the choice. However, RCTs can be unpractical and infeasible in certain clinical conditions. Therefore, starting from analyzing the observational data becomes an alternative. With the improvement of data collection and advancement in DRL technologies, we see great potentials in DRL-based decision support system to optimize treatment recommendations.\\

\paragraph{Technical Significance}
This survey paper summarizes and discusses major types of DRL algorithms that have been applied to clinical applications to provide clinical decision support. Furthermore, we discuss the trade-offs and assumptions for these DRL algorithms. This paper aims to serve as a guideline to assist our audience to pick the appropriate DRL models for their particular clinical applications. To the best of our knowledge, this is the first survey paper on DRL for treatment recommendation or clinical decision support.

\paragraph{Clinical Relevance}
DRL is proven to achieve the human-level capacity for learning complex sequential decisions in specific domains, such as video game, board game, and autonomous control. While in healthcare, DRL has not been widely applied for clinical applications yet. In this paper, we survey major DRL algorithms that provide sequential decision support in the clinical domain. We believe that learning from the vast amount of collected electronic health record (EHR) data, DRL is capable of extracting and summarizing the knowledge and experience needed to optimize the treatments for new patients. DRL also has the potential to expand our understanding of the current clinical system by automatically exploring various treatment options and estimate their possible outcomes.

\paragraph{Structure}
This paper surveys the reported cases on the applications of DRL algorithms for clinical decision support. In part 2, we first introduce the basic concepts of RL and DRL. Then we summarize the main types of DRL algorithms. In part 3, we present a few clinical applications that use various DRL algorithms. In part 4, we discuss how to choose among various DRL algorithms. Finally, in part 5, we investigate the challenges and the remedies of using DRL for clinical decision support.

\section{Fundamentals of Reinforcement Learning}
Reinforcement learning (RL)\citep{sutton1998introduction} is a goal-oriented learning tool wherein an agent or a decision maker learns a policy to optimize a long-term reward by interacting with the environment. At each step, an RL agent gets evaluative feedback about the performance of its action, allowing it to improve the performance of subsequent actions\citep{kiumarsi2018optimal}. Mathematically, this sequential decision-making process is called the Markov Decision Process (MDP)\citep{howard1960dynamic}.

\subsection{MDP Formulation}
A Markov decision process is defined by 4 major components:\\
\begin{itemize}
    \item A state space $S$: at each time t, the environment is in state $s_t \in S$
    \item An action space $A$: at each time $t$, the agent takes action $a_t \in A$, which influence the next state, $s_{t+1}$\\
    \item A transition function $P(s_{t+1} \given s_t, a_t)$: the probability of the next state given the current state and action, which represent an environment for the agent to interact.
    \item A reward function $r(s_t, a_t) \in \mathbb{R}$: the observed feedback given the state-action pair $(s_t, a_t)$.\\
\end{itemize}
In a clinical context, we consider an agent as a clinician. We can think the state as the wellbeing/condition of a patient. The state of the patients can depend on his demographics (e.g., age, gender, ethnicity, etc.), longitudinal and physiological measurements (e.g., lab test, vital signs, medical image reports, etc.) and some other clinical features. An action is a treatment that clinicians act to the patient (e.g., prescription of certain medication, ordering of surgical procedures, etc.). The transition function $P(s_{t+1} \given s_t, a_t)$ can be view as the patient's own biological system, that given the current wellbeing and the intervention, the patient will enter the next time step $s_{t+1}$. If the wellbeing is improved, we assign a reward to the agent, and we penalize the agent if the patient's condition gets worse or stay stagnant after the intervention.\\

The objective of the reinforcement learning agent\citep{sutton2018reinforcement} is defined as the expected total reward along with a trajectory $\tau$ that follow the distribution $p(\tau)$, i.e., a mapping from state to action $\pi(a \given s)$, that maximizes the expected accumulated reward.\\
\begin{equation}
\pi^* = arg \max\underbrace{E_{\tau \sim p(\tau)}[\sum\limits_{t}\gamma^tr(s_t, a_t)]}\limits_{J}\tag{1}
\end{equation}
where $p(\tau)$ is the probability distribution over state-action trajectories, and $\gamma$ is a discounted factor.\\

The expected long term reward noted as $J$ in equation (1), is the objective that we want to maximize w.r.t. a policy $\pi(a \given s)$. The optimal policy $\pi^*$ is the one that maximizes the objective $J$.\\

To evaluate how 'good or bad' a state or an action is, we define Q-function and value-function in equation (2) and (3) respectively.
\begin{equation}
    Q^\pi(s_t, a_t) = \sum_{t^\prime=t}^{T}E_{\pi}[r(s_{t^\prime}, a_{t^\prime})\given s_t, a_t]\tag{2}
\end{equation}

\begin{equation}
    V^\pi(s_t) = E_{a_t \sim \pi(a_t, s_t)}[Q^\pi(s_t, a_t)]\tag{3}
\end{equation}

Q-function $Q^\pi(s_t, a_t)$ is the conditional expectation of cumulative reward given the current state-action pair $(s_t, a_t)$, whereas the value  $V^\pi(s_t)$ is the expectation of Q-value at $(s_t, a_t)$ and take action $a_t$ according to the policy $\pi$.\\

If we know the actual Q-function or value-function for all possible states and actions in an environment, an optimal policy $\pi^*$ will be easy to derive from equation (1). An agent at each state $s_t$ will pick an action if it maximizes the Q-function or the value-function.

\subsection{Deep Reinforcement Learning}
Deep Reinforcement Learning (DRL) adopts deep neural networks (DNN)\citep{cirecsan2012multi} to replace  tabular representations for $Q(s_t, a_t)$, $V(s_t)$ or $\pi$ in traditional RL. For complex environments with a large number of states, tabular representations are not feasible due to the large-scale storage and high computation requirements (Curse of Dimensionality).\\

Different DRL algorithms use DNN in various ways. For instance, in policy gradient RL\citep{sutton2000policy}, DNN can be used to optimize a policy $\pi(\theta)$, where $\theta$ is the parameters (weights) in DNN. The input to the DNN is the current state, and the output is an action. By taking the gradient of DNN, at each time step, the parameters $\theta$ in the DNN are updated and thus the policy is improved.\\

In value-based RL\citep{kaelbling1996reinforcement}, DNN can be used to approximate the value function or the Q-function. For the value function, DNN take the current state $(s_t)$ as the input and the output is the approximated value $\hat{V}^\pi (s_t)$ given a policy $\pi$. whereas for Q-function approximation, DNN take the state-action pair $(s_t, a_t)$ as the input and output is the  $\hat{Q}_\phi(s_t, a_t)$ value, where $\phi$ is the parameter in DNN for value-based RL. In model-based RL\citep{doya2002multiple}, DNN is used to approximate the transition function $p_\phi(s_{t+1} \given s_t, a_t)$.\\

We will discuss the architecture for different DRL in details in section 2.4. In order to have a clearer understanding of DRL, let us first discuss the key steps in RL algorithms, then we will introduce main types of RL algorithms, including policy gradient RL, valued-based RL, actor-critic RL\citep{grondman2012survey}, model-based RL and some other extensions of RL algorithms. (From this point onward, all the RL algorithms we discuss in this paper refer to RL with DNN architecture (DRL). We will use the term 'RL' and 'DRL' interchangeably, while they both refer to Deep RL.)

\subsection{Key Steps in Reinforcement Learning}
Most RL algorithms can be broken down into 3 steps:\\
\textbf{Step 1. Sample collection}\\
\textbf{Step 2. Evaluate objective}\\
\textbf{Step 3. Improve policy}\\

We take policy gradient RL as an example. In step 1, the RL algorithm interacts with the environment and generate sample (state-action pairs) by following an unlearned policy (The initialization can be random). The set of state-action pairs are sequential and form a trajectory $\tau$. The RL algorithm can generate multiple trajectories in step 1. In the next step, the RL algorithm will try to evaluate how good or bad are those collected trajectories by calculating the objective function $J$ (cumulative reward). Then in step 3, the RL algorithm will try to maximize the objective $J$ by increasing the chance of visiting those highly rewarded trajectories and reducing the chance of visiting low rewarded trajectories, and this is called policy update. The agent will then go back to step 1 to generate new samples according to the update policy and will repeat step 2 and 3 to update the policy. In this way, the policy is improved over time by 'trial and error'.\\

Most RL algorithm follows this three-step structure, but the three steps are not equally important in different RL algorithms. So, which RL algorithm works the best for an application is a context-dependent question. We will discuss this question in more details in section 4. For now, let us go through the cost in the three steps briefly. In step 1, if we collect data samples from a real physical system, then data generation would be expensive because the data collection is in real time. In contrast, if we use simulators to generate samples, like those in Atari games\citep{mnih2013playing}, then samples are much easier to obtain and step 1 would probably not be a concern in this case.\\

For step 2, if the application uses policy gradient to improve policy, where the policy can be improved by trial and error, then the objective $J$ will be the sum of a few rewards from those trials. However, for model-based RL, if the model is fitted using DNN, then fitting the model will be expensive that we need to make sure the fitted model converges and close to the actual model.\\

In step 3, policy gradient RL will only require us to compute the gradient and apply the gradient to the policy. It is still relatively simple. However, for model-based RL with DNN, we will need to do the backpropagation for a vast number of time steps to improve the policy.\\

When it comes to real-world applications, especially in clinical applications, data generation is often a big concern because it is in real time, and many of the observational data (from Electronic Health Records) are private that we do not readily have access. We also do not wish to estimate values by trials and error, because it is often unethical and costly. Therefore, when we choose the RL algorithms, we need to be aware of those constraints.\\

In this paper, we will first focus on discussing the main types of RL algorithms that were applied for clinical decision support, while only briefly cover one RL algorithm which is rarely applied in clinical applications. We will also discuss the underlying reason why specific algorithms are not so "popular" compared to others in the healthcare domain.

\subsection{Types of Reinforcement Learning Algorithms}
\paragraph{Policy Gradient Reinforcement Learning}

Under section 2.3, we discussed the three-step structure of general RL algorithm. Policy Gradient RL follows this three-step structure. In step 1, the policy gradient RL first rolls out a random policy $\pi(\theta)$ to the environment and generate trajectories under the policy. The environment is the transition function that does not need to specify in this setting. Because we are not interested to learn the exact transition function in policy gradient RL. After a few trials, we collect a set of trajectories, and we come to step 2 to evaluate the reward $J(\theta)$. It can be approximated by samples from trajectories that are collected in step 1. $J(\theta) \approx \frac{1}{N}\sum\limits_{i}\sum\limits_{t}r(s_{i,t}, a_{i,t})$
In step 3, we'll apply gradient to the expected long term reward $J(\theta)$ and update the policy $\pi_\theta$, where new $\theta^\prime$ is set to $\theta + \alpha \nabla_\theta J(\theta)$.Then we need to sample new state-action pairs from the updated policy $\theta^\prime$ in step 1 and do the optimization of policy step by step. We call this on-policy\citep{krstic1995nonlinear}, because every time the policy changes, we need to sample new data under the new policy. In contrast to on-policy, off-policy applies when we can improve the policy without generating new samples from the policy. Therefore, policy gradient RL is an on-policy RL algorithm.\\

In policy gradient RL, DNN is used to construct a policy in step 3, where the input to the DNN is state and output is an action. Figure 1 is the MDP diagram of policy gradient RL. By taking a gradient of $J(\theta)$ and update the weights in DNN, the policy is learned accordingly. In the clinical context, policy gradient RL is not as 'popular' compared to other RL algorithms. The underlying reason for this may lie on the fact that it is an on-policy algorithm that needs to collect data iteratively based on a new policy. The algorithm is learned by 'trial and error.' Most clinical applications cannot afford the cost of collecting real-time clinical data. For instance, to learn the optimal clinical decisions of medication dosage for sepsis patients, it would be unethical to do trial and error and will also be time-consuming. However, policy gradient RL is still popular under other domains, such as robot control and computer board game, where the environment is a simulator that can afford for the trial and error.
\begin{figure}[htbp]
    \centering 
    \includegraphics[width=4in]{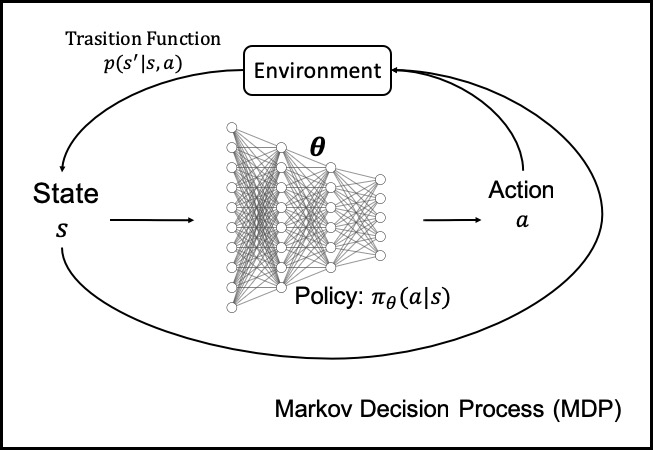} 
    \caption{MDP for policy gradient RL algorithm}
    %\label{fig:example} 
\end{figure} 

\paragraph{Value-based Reinforcement Learning}

Value-based RL tries to estimate the value function or Q-function for certain state and action. The Q-function in equation (2) is also known as bellman equation. Value-based RL tries to evaluate either value function $V^\pi(s_t)$ or Q-function $Q^\pi(s_t, a_t)$ from the conditional cumulative reward using DNN in step 2. While for value function, the input to the DNN is state $s_t$, and output is $V^\pi(s_t)$; For Q-function, the input to the DNN is state-action pair $(s_t, a_t)$, and output is $Q^\pi(s_t, a_t)$. The DNN is trained to optimize the Mean Squared Error (MSE) given as  $L(\phi)= \frac{1}{2}\norm{\hat{V}_\phi^\pi(s_t)-y_{t+1}}^2$, where $y_{t+1}$ is called target value. $y_{t+1} = \max\limits_{a_{t+1}}\gamma Q^\pi(s_{t+1}, a_{t+1})$ for value function and $y_{t+1} = r(s_t, a_t)+ \gamma \max\limits_{a_{t+1}}Q^\pi(s_{t+1}, a_{t+1})$ for Q-function. This optimization can be done with various Gradient Descent methods in deep learning, such as Stochastic Gradient Descent (SGD)\citep{bottou2010large}.  So DNN is implemented in step 2 of the RL algorithm since it's evaluating the objective $J(\theta)$ that is estimated by Q-function or value function. In step 3, value-based RL update the policy $\pi$ to $\pi^\prime$ only if the action taken from $\pi^\prime$ resulted in the maximum for Q-function.\\
\begin{equation}
    \pi^\prime(a_t, s_t)= 
\begin{cases}
1,& \text{if } a_t = arg \max_{a_t}Q^\pi(s_t, a_t),\\
0,& \text{otherwise}
\end{cases}
\tag{4}
\end{equation}

These two algorithms are called \textbf{Fitted Value Iteration (FVI)}\citep{munos2008finite} and \textbf{Fitted Q Iteration (FQI)}\citep{riedmiller2005neural} respectively. A special case for FQI is called \textbf{Q-learning}\citep{watkins1992q}, wherein step 1 we only take one tuple of sample $(s_t, a_t, s_{t+1}, r_t)$ and use that for Q-function approximation. Wewe use one gradient step to do parameter update, then go back to step 1 to collect one more tuple of the new sample and do the optimization iteratively. The term 'Q-learning' also sometimes refers to general value-based RL in some literature.\citep{arulkumaran2017brief}\\

 FVI, FQI, and Q-learning work well for off-policy samples and do not have a high-variance as in policy gradient RL\citep{arulkumaran2017brief}. However, the limitation is that they often not converge for non-linear function approximation (such as DNN). Besides, the samples collected from Q-learning are temporal sequential; thus samples are highly correlated\citep{arulkumaran2017brief}. Then the network might only find the optimal local solution. Another issue with Q-learning is that its target value $y_{t+1}$ is estimated using one tuple sample and iterate over multiple one-step estimations. It makes the target value very unstable.\\
 
\cite{lin1992self} mitigated the problem of highly correlated samples using a new component named \textbf{'replay buffer'} with Q-learning. \textbf{Q-learning with replay buffer} sample a batch of tuples $(s_t, a_t, s_{t+1}, r_t)$ and use the batch to do a one-step gradient for parameter update and go back to the buffer to collect another batch of tuples. In this way, samples are no longer correlated, and multiple samples in the batch also ensure low gradient variance. To mitigate the problem of the unstable target value,  \textbf{'target network'} is introduced by \cite{mnih2016asynchronous}, where the target value $y_{t+1}$ is estimated by a fixed value and only gets updated after a few iterations of learning. Combining the two components, 'replay buffer' and 'target network' with Q-learning, \cite{mnih2015human} defined a new RL algorithm \textbf{Deep Q-Network (DQN)}.\\
 
 However, DQN is also not perfect. It has the same limitation as Q-learning that it often overestimate Q-function \citep{thrun1993issues}. The overestimation is due to '$\max$' term in the estimated target value function $y_{t+1} = r(s_t, a_t)+ \gamma \max\limits_{a_{t+1}}Q^\pi(s_{t+1}, a_{t+1})$, which is affected by noise during data collection\citep{van2011insights}. One possible method to mitigate this issue is to use \textbf{Double Q-Learning (Double DQN)} where \citep{hasselt2010double} implemented two DNNs to learn two Q-functions.  They used one of them to evaluate the Q-value and let the other one choose the action. The goal of having two Q-functions is to de-correlate the noise in both action and Q-function.\\
 
 Value-based RL is commonly applied in clinical applications, and we will see more examples from section 3 Clinical applications on DRL.

\paragraph{Actor-Critic Reinforcement Learning}

Actor-critic RL\citep{heess2015learning} is a combination of policy gradient RL and value-based RL. In step 1, we sample state-action pairs from policy just like those on the policy gradient RL, but this time, the state-action pairs are generated by a DNN (actor) that act according to the policy $\pi_\theta$. Then, in step 2, we fit the value $\hat{V}_\phi^\pi(s_t) = \sum_{i=t}^{T}E_{\pi_\theta}[r(s_i, a_i)\given s_t]$ to sampled reward sums using another DNN (critic), the input to DNN is state s, and the output is  $\hat{V}^\pi(s)$. We evaluate a new term named 'advantage'\citep{baird1993advantage,harmon1996multi} $\hat{A}^\pi(s_i, a_i)$, which is defined as $\hat{A}^\pi(s_i, a_i)= r(s_i, a_i)+ \hat{V}_\phi^\pi(s^\prime) - \hat{V}_\phi^\pi(s)$. where $\hat{V}_\phi^\pi(s^\prime)$ is the estimated value for the next state $s^\prime$. Advantage function tells us how much better is that action than the average action in the state according to the estimated value function. Then in step 3, we take the gradient of the cumulative advantage $\hat{A}^\pi(s_i, a_i)$ along the trajectory and update $\theta$ in the actor DNN to learn a better policy. The critic DNN can be optimized by supervised regression, $L(\phi)= \frac{1}{2}\norm{\hat{V}_\phi^\pi(s_t)-y_t}^2$, where $y_t$ can be estimated by either Monte Carlo estimation of reward along the trajectory, or it can be estimated by bootstrap method. The Actor-critic RL is an off-policy algorithm, but an alternative version can also be an on-policy algorithm. The only difference is in step 1, instead of collecting a batch of trajectories, we only collect one trajectory and update the policy to generate new samples from the updated policy. Again, the on-policy will not be suitable to implement in real-time clinical applications; thus one of the applications discussed in this paper\citep{wang2018supervised} utilized the off-policy actor-critic RL algorithm.

\paragraph{Model-based Reinforcement Learning}

All the above discussed RL algorithms are all model-free RL, and in model-free RL we assume we do not know about the exact transition function $p(s_{t+1} \given s_t, a_t)$. So given the current state and action pair, we do not know what the real next state is. The model-free RL does not attempt to learn the transition function explicitly but bypass it by sampling from the environment. Knowing the right transition function or the environment will always be helpful. Moreover, in certain situations, we do know about the transition function, such as a simple board game where we design the rules ourselves. For clinical applications, most of the time we are not sure about the exact transition function, but we do know a bit about the dynamics of the environment. For instance, clinicians generally know that after treating a sick patient with appropriate medication dosage, the patient will gradually recover from a sick state to a healthy state. Even if we do not know the full picture of the environment, we can still propose several models to estimate the real transition function (environment) and optimize it from there. It is called model-based RL\citep{doya2002multiple}. There are various models we can use to estimate the transition function, such as Gaussian process(GP)\citep{deisenroth2011pilco,rasmussen2003gaussian}, DNN, Gaussian mixture models (GMM)\citep{chernova2007confidence}, and so on. With DNN model-based RL, the input to the DNN is state-action pair $(s_t, a_t)$ and output is $s_{t+1}$. The DNN is implemented in step 2 for the transition function.
In contrast to DNN as the model for the environment, GP is very data efficient. GP can produce reasonably decent predictions of the next state using few data samples. It is useful in the clinical context since most of the clinical application suffers from the data deficiency problem. However, the limitation for GP is that it has troubles when the actual transition function is non-smooth. Moreover, GP can be slow if the number of samples is vast and in high dimensional space. It is the opposite of DNN, wherein DNN larger the number of samples, the more accurate prediction in general. So in clinical context when the input state is medical images (very high dimensional), DNN will be more suitable for model-based RL than GP.

\paragraph{Other Extensions of Reinforcement Learning}
\begin{itemize}
    
    \item \textit{Hierarchical Reinforcement Learning}\\
    
    When the learning task is massive, and we have seen several different state-action spaces in RL with several sub-optimal policies, it is intuitive to sequence the sub-space and try to obtain the optimal policy for global space. Hierarchical RL\citep{kulkarni2016hierarchical} generally contains a two-level structure. The lower level is just like the policies we trained in general RL algorithms that try to suggest an action given $(s_t, a_t)$. In the same time, there exists a higher level network where the 'meta-policy' trains to select which of these lower level policies to apply over a trajectory. Hierarchical RL has the advantage of learning faster global optimal policy compared to policies randomly initiated, and it transfers the knowledge learned from past tasks from lower level policies. In the clinical setting, the state-action space can be huge due to the complex behavior of human interactions. Therefore, applying the Hierarchical RL is a very natural choice for clinical applications. However, the architecture for Hierarchical RL is more complicated to train, and inappropriate transfer learning can result in 'negative transfer'\citep{pan2010survey} where the final policy is not necessarily better than lower level policies. 
    
    \item \textit{Recurrent Reinforcement Learning}\\
    
    A fundamental limiting assumption of Markov decision processes is the Markov property (full observation of MDP), which is rarely satisfied in real-world problems. In medical applications, it is unlikely that a patient’s full clinical state will be measured. It is known as a Partially Observable Markov Decision Process (POMDP) problem\citep{kaelbling1998planning}. A POMDP has a 4-tuple $(S, A, R, O)$, where $O$ are observations. Classic DQN is only useful if the observations reflect the underlying state.\citep{hausknecht2015deep}. \cite{hausknecht2015deep} proposed an extension to DQN network to deal with POMDP problem, where the first fully connected layer of DQN is replaced with Long Short Term Memory (LSTM)\citep{hochreiter1997long}. This new RL algorithm is called Deep Recurrent Q-Network (DRQN)\citep{hausknecht2015deep}. Their algorithm showed that it could integrate information successfully through time and could replicate DQN’s performance on standard Atari games with a setting of POMDP for the game screen.
    
    \item \textit{Inverse Reinforcement Learning}\\
    
    To learn most of the standard RL algorithm in clinical applications, we would design the reward function by hand, but we do not know what the real reward is. This reward design is very vulnerable to if it is misspecified. Inverse RL is an algorithm that we can infer the correct reward function from expert demonstrations without having to program the reward by hand.\citep{ghavamzadeh2015bayesian,abbeel2004apprenticeship,ng2000algorithms} Alternative to Inverse RL is to learn them directly from the behaviors from experts, and this often refers to imitation learning\citep{schaal1999imitation}. However, one limitation for imitation learning is that the experts may have different capabilities and prone to be imperfect\citep{wang2018supervised}; learning from experts may only result in sub-optimal policies. So generally in Inverse RL,  we are given state-action space and a sub-optimal policy in a dynamic model. The goal of Inverse RL is to recover the right reward function. Then we can use the learned reward function to get a new policy which is better than the sub-optimal policies. The reward can be learned by DNN where the input is state-action pairs $(s, a)$ produced by a sub-optimal policy $\pi^\#$, then the output of the DNN is a reward $r_\Phi(s, a)$, where $\Phi$ is the parameters to the DNN that we would learn through backpropagation.     
    Later after we obtained $r_\Phi(s, a)$, we can use the new reward function to plan for better policy and hopefully the optimal policy $\pi^*$.  In the clinical context, it is important to reason about what the clinicians are trying to achieve and what they think is essential.
\end{itemize}

\section{Deep Reinforcement Learning for Clinical Applications}
Recent research\citep{prasad2017reinforcement, nemati2016optimal, tseng2017deep} have demonstrated that DRL can be used to provide treatment recommendation and clinical decision support for various applications with different types of data sources, ranging from EHR, online disease database, and genetic data, and so on. The applications include medication/fluid choice, dosage for patients with acute or chronic disease, settings and duration of mechanical ventilation, and constructing clinical motifs from clinical notes. Table 1 summarizes all the clinical application papers discussed in this survey, in particular highlighting the distinct sub-type of DRL methods and different clinical problems. As discussed in the earlier section, the policy gradient RL is an on-line algorithm that does not fit with most clinical applications. Therefore, value-based RL algorithms were popular in clinical applications.\\

For each application in this section, we will investigate what RL algorithms the author used, the formulation of context-based MDP  $(S, A, R)$, and finally the performance of the RL. We first discuss the applications that used value-based RL, followed by actor-critic RL, model-based RL, hierarchical RL, recurrent RL and finally inverse RL. For value-based RL, we further divide the application based on the sub-types of value-based RL algorithms, which contain Fitted Q Iteration, DQN, and double DQN. 

\subsection{Value-based Reinforcement Learning}

\textbf{{\small Fitted Q Iteration}}\\

\underline{Weaning of Mechanical Ventilation}\\

Weaning patients in the intensive care unit from mechanical ventilation (MV) is often haphazard and inefficient. With this regard, \cite{prasad2017reinforcement} used an off-policy Fitted Q Iteration (FQI) algorithms to determine ICU strategies for the MV administration, they aimed to develop a decision support tool that could leverage available patient information in the data-rich ICU setting to alert clinicians when a patient was ready for initiation of weaning, and to recommend a personalized treatment protocol. They used MIMIC-III database to extract 8860 admissions from 8182 unique adult patients undergoing invasive ventilation for more than 24 hours.\\
They included features like patients' demographic characteristics, pre-existing conditions,  co-morbidities and time-varying vital signs. They preprocessed the lab measurements and vitals using Gaussian Process (GP) to impute missing values; this could ensure more precise policy estimation. The state $s_t$ was a 32-dimensional feature vector. The action was designed as a 2-dimensional vector with the 1st dimension of on/off MV and the 2nd dimension for four different levels of dosage for sedation. Reward $r_t$ with each state encompassed both time-into-ventilation and physiological stability, whereas reward would increase with stable vitals and successful extubation, but would penalize on adverse events (reintubation) and additional ventilator hour. Comparing to the actual policy implemented at the Hospital of University of Pennsylvania, HUP, their learned policy matched 85\% of actual policy.\\

\underline{Optimal Heparin Dosing in ICU}\\

Mismanagement of certain drugs can drive up costs by unnecessarily extending hospital length of stay, and place patients at risk. Unfractionated Heparin(UH) is one example of such drugs that overdosing leads to increased risk of bleeding and underdosing leads to increased risk of clot formation. RL is particularly well-suited for the medication dosing problem given the sequential nature of clinical treatment. \cite{nemati2016optimal} trained an RL medication dosing agent to learn a dosing policy that maximized the overall fraction of time given patient stays within his/her therapeutic activated partial thromboplastin time (aPTT). They used the MIMIC-II database and extracted 4470 patients who received an intravenous heparin infusion at some point during their ICU stay with a time window of 48 hours. Variables included demographics, lab measurements and severity scores(Glasgow Coma Score(GCS), daily Sequential Organ Failure Assessment(SOFA)score). A state was constructed using the features and estimated by discriminative hidden Markov model (DHMM)\citep{kapadia1998discriminative}. Whereas the discrete action was using discretized heparin values from six quantile intervals, and the reward was designed according to the aPPT feedback. Given the feedback, a decision was made to increase, decrease or maintain the heparin dosage until the next aPTT measure. $r_t=\frac{2}{1+e^{-(aPTT_t-60)}} - \frac{2}{1+e^{-(aPTT_t-100)}}-1$. This function assigned a maximal reward of one when a patient's aPTT value was within the therapeutic window which rapidly diminished towards a minimal reward of -1 as the distance from the therapeutic window increases.\\
The performance of the optimal policy was tested by comparing the accumulated reward from a clinician and the trained RL agent. On average and consistently over time, the RL algorithm resulted in the best long-term performance by following the recommendations of the agent.\\

\textbf{\small{Deep Q Network}}\\

\underline{Extract Clinical Concepts from Free Clinical Text}\\

Extracting relevant clinical concepts from a free clinical text is a critical first step for diagnosis inference. \cite{ling2017diagnostic} proposed to use DQN to learns clinical concepts from external evidence (Wikipedia: Signs and Symptoms Section and MayoClinic: Symptoms section). They used TREC CDS dataset\citep{simpson2014overview} to conducted their experiments. This dataset contains 30 topics, where each topic is a medical free-text that described a patient's situation with a diagnosis. MetaMap\citep{aronson2006metamap} extracted clinical concepts from TREC CD5 and external articles. The state contained two vectors: current clinical concepts, as well as candidate concepts from external articles: the more similar between the two vectors, the higher state value, would be.  The action space was discrete that included action to accept or reject candidate concepts. The reward was designed in a way such that it was high when candidate concepts were more relevant than current concepts for a patient's diagnosis. DQN was trained to optimize the reward function that measured the accuracy of the candidate clinical concepts. Their preliminary experiments on the TREC CDS dataset demonstrated the effectiveness of DQN over various non-reinforcement learning based baselines.\\

\underline{Symptom Checking 1.0}\\

To facilitate self-diagnosis, \cite{tang2016inquire} proposed symptom checking  systems. A symptom checker first asked a sequence of questions regarding the condition of a patient. Patients would then provide a sequence of answers according to the questions. Then the symptom checker tried to make a diagnosis based on the Q\&A. Tang et al. proposed an RL based ensemble model to train this symptom checker. They implemented 11 DQN models, while each model represented an anatomical part of a human body, such as head, neck, arm, etc. Those model were trained independently of each other. For each anatomical model, the state s was one-hot encoded based on a symptom. (i.e., if the symptom were a headache, then only the anatomical model representing head would have $s=1$, while other models would have $s =0$). The action was discrete and had two types: inquiry and diagnosis. If the maximum Q-value corresponded to the action of inquiry, the symptom checker would continue to ask the next question. If the maximum Q-value corresponded to the action of diagnosis. The symptom checker would give a diagnosis and terminate. The reward was designed as a scalar. The agent would receive the reward when it could correctly predict the disease with a limited number of inquiries. They applied the DQN algorithms on a simulated disease dataset, and the result showed that symptom checker could imitate the behavior of inquiry and diagnosis as those performed by doctors.\\

\textbf{\small{Double DQN}}\\

\underline{Sepsis Treatment 1.0}\\

\cite{raghu2017continuous}. was one of the first ones to directly discuss the application of DRL to healthcare problems. They used the Sepsis subset of the MIMIC-III dataset and chose to define the action space as consisting of Vasopressors and IV fluid. They grouped drugs doses into four bins consisting of varying amounts of each drug. Q-values were frequently overestimated in practice, leading to incorrect predictions and poor policies. Thus the authors solved this problem with a Double DQN\citep{wang2015dueling}. They also employed dueling deep Q Network to separate value and advantage streams, where the value represented the quality of the current state, and the advantage represented the quality of the chosen action. The reward function was clinically motivated based on the SOFA score which measures organ failure. They demonstrated that using continuous state space modeling, the found policies could reduce patient mortality in the hospital by 1.8\% to 3.6\%.\\

\subsection{Actor-critic Reinforcement Learning}
\underline{Optimal medical prescription in ICU}\\

\cite{wang2018supervised} adopted the Actor-critic RL algorithms to find optimal medical prescriptions to patients with various diseases. They experimented with MIMIC-III database and extracted 22,865 admissions. Features used for state construction included demographics, vitals, lab results, etc. Action space was 180 distinct ATC codes. Wang et al. not just implemented classic actor-critic RL. Instead, they combined both RL with Supervised learning (SL) in the actor-network. The objective function $J(\theta)$ was evaluated as the linear combination of objective function for both RL and SL in the equation: $J(\theta)=(1-\epsilon)J_{RL}(\theta) + \epsilon(-J_{SL}(\theta))$, where $\epsilon$ was a hyperparameters that ranges from 0 to 1, and it used to balance the RL and SL.  $J_{RL}(\theta)$ was objective in actor-network, while $J_{SL}(\theta)$  was evaluated as the cross-entropy loss between the predicted treatment and prescriptions given by the doctor.\\
They applied gradient ascent to the objective function w.r.t. $\theta$ in the actor-network,  and tried different $\epsilon$ value for RL-SL balancing. Besides, they also incorporated an LSTM network to improve performance in the partial observed MDP (POMDP). The state $s$ was replaced by summarizing the entire historical observations with $c_t = f(o_1, o_2, \dots, o_t)$ and $c_t$ was used as state for both actor and critic networks. Their experiments showed that the proposed network could detect good medication treatment automatically.\\

\subsection{Model-based Reinforcement Learning}
\underline{Radiation dose fraction for lung cancer}\\

\cite{tseng2017deep} implemented model-based RL to train a dose escalation policy for patients received radiotherapy with lung cancer. They included 114 patients in the RL design and first trained a DNN to estimate the transition function $p(s_{t+1}\given s_t, a_t)$. The loss function for the DNN model aimed to minimize the difference between the expectation of Q-values from the estimated trajectory with the observed values. After the construction of the transition function, Tseng et al. applied DQN to learn an optimal policy for the dose in thoracic irradiation that trade-off between the local control (LC) and the risk for radiation-induced pneumonitis (RP). The reward for the network was designed as a trade-off between encouraging improved LC and attempting to suppress RP. State of DQN was defined as a combination of 9 features, including cytokines, PET radiomics, and dose features. The action was designed as the dose per fraction.\\
Since the construction of the transition function would require a large amount of data. The authors implemented drop-off in the DNN in transition function to avoid overfitting. Besides, the authors implemented a Generative Adversarial Networks (GAN)\citep{goodfellow2014generative} to simulate more data to mitigate the data deficiency problem. The simulated data from GAN was also fed to the transition function DNN to training.  The proposed (Model-based RL) network showed a promising result that it could suggest similar treatment dose compared with clinicians.\\

\subsection{Hierarchical Reinforcement Learning}
\underline{Symptom Checking 2.0}\\

The main idea from \cite{kao2018context}'s work was to imitate a group of doctors with different expertise who jointly diagnose a patient. Since a patient could only accept an inquiry from one doctor at a time, a meta-policy was required to appoint doctors, in turn, to inquire to the patient. The meta-policy came from a higher level network. At each step, the meta-policy was responsible for appointing an anatomical-part model to perform a symptom inquiry for a disease prediction.\\
In Kao et al's paper, the first hierarchy level was a master agent M. The master M possesses its action space $A_M$ and policy $\pi_M$. In this level, the action space $A_M$ equaled $P$, the set of anatomical parts. At step t, the master agent entered state $s_t$, and it picked an action $a_{M_t}$ from $A_M$ according to its policy $\pi_M$. The second level of hierarchy consists of anatomical models $m_p$. If the master acted $a^M$, the task would be delegated to the anatomical model $m_p = m_{a^M}$. Once the model $m_p$ was selected, the actual action $a_t \in A$ was then performed according to the policy of $m_p$, denoted as $\pi_{m_p}$.\\
Based on this structure, Kao et al. trained an online symptom checkers on simulated data from SymCAT symptom disease database for self-diagnosis of health-related ailments. It was the improved version of Symptom Checking 1.0, where Kao et al. added another layer of DQN on top of anatomical models in Symptom Checking 1.0 as a master agent.  Both the anatomical models and the master model applied DQN to pick the action that maximized the Q-value. Their result showed that the proposed Hierarchical RL algorithm significantly improved the accuracy
of symptom checking over traditional systems.\\

\subsection{Recurrent Reinforcement Learning}
\underline{Sepsis Treatment 2.0}\\

\cite{futoma2018learning} proposed a new extension to DRQN architecture with a multi-output Gaussian process to train an agent to learn the optimal treatment for sepsis patients. They collected data from private database in the Duke University health system with 9,255 sepsis patients and their 165 features (including demographics, longitudinal physiological variables, medication, etc. ), and followed up in 30 days. Actions were discrete values consisting of 3 common treatment for sepsis patients: antibiotics, vasopressors, and IV fluids. The reward was sparsely coded. The reward at every non-terminal time point was 0. A reward was +10 at the end of a trajectory if the patient survives; and -10 if the patient dead. They investigated the effect of replacing fully connected neural network layers with LSTM layers in the DQN architecture. The optimized policy for sepsis treatment could reduce patient mortality by as much as 8.2\% from an overall baseline mortality rate of 13.3\%.

\subsection{Inverse Reinforcement Learning}
\underline{Diabetes Treatment}\\

In previous papers, the reward function was approximated by the heuristic. However, the appropriateness of the reward function could not be validated. In recent studies on RL, inverse RL has been proposed to estimate the reward function of experts from their behavior data. There have been some papers that focus on Inverse RL\citep{ng2000algorithms, abbeel2004apprenticeship}. However, to the best of our knowledge, there was not any paper that implemented DNN based Inverse RL algorithm for clinical decision support. For non-DNN based Inverse RL, \cite{asoh2013application} implemented an Inverse RL in the Bayesian framework and used Markov chain Monte Carlo sampling\citep{ghavamzadeh2015bayesian} to learn the reward function in a clinical context. They applied the Inverse RL to learn the reward function for diabetic treatments with a uniform prior. The drug prescription data was private with the University of Tokyo Hospital. The state was discrete and defined as the severity of diabetes ('Normal', 'Medium', 'Severe'). They used MCMC-sampling and derived the reward for the 3 states as $r = (0.01, 0.98, 0.01)$. The reward showed 'medium'-level diabetes patients have the highest value. It seemed to contradict with the current clinical understanding that the reward should have the highest value for 'normal'-level diabetic patients. Asoh et al. explained that 65\% of diabetic patients from their database were already in the 'medium' condition. Therefore, keeping the patients in the 'medium' state might be the best effort from clinicians.\\

Despite very few clinical application implemented Inverse RL, we believe that inverse RL is a valuable topic, and it will benefit the applications in the clinical context. We are not only optimizing the policy by mimicking experts behavior, but we are also very keen to train the policy such that it can automatically figure out what are the treatments that clinicians think important.

\section{How to Choose among Reinforcement Learning Algorithms}
There is no unique answer to this question. Choice of RL algorithm will depend on the actual application. Below is a list of trade-offs and assumptions to consider: \\

\textbf{Sample efficiency}: Sample efficiency refers to how many samples are required to make the algorithm converge. If the algorithm is on-policy, such as policy gradient RL, then it would take more samples. In contrast, value-based RL and model-based RL are off-policy algorithms, so that fewer samples are required for training. Actor-critic RL algorithm is in between value-based RL and policy gradient RL. Given different sample efficiency, it does not mean we should always choose the one that requires fewer samples. For specific applications where samples are easily generated (i.e., symptom checker 1.0 used simulator to generate data), the wall-clock time for model training may be more important than the number of samples required. In such cases, on-policy RL algorithms might be preferred because they are generally faster to converge than off-policy RL algorithms.\\

\textbf{Convergence}: Policy gradient performs gradient ascent on the objective function, and it is guaranteed for convergence. Value-based RL minimizes the "Bellman error" of fit, but in the worst case, it is not guaranteed to converge to anything in the nonlinear cases. Whereas for model-based RL algorithm, the model minimizes the error of fit and the model is guaranteed to converge. However, a better model is not equivalent to better policy.\\

\textbf{Episodic/infinite horizon}: Episodic means the there is an end-point for a state-action trajectory. For instance, in the disease symptom checker application,  an agent continuously searches for the symptom, an episode is over when the agent found the disease. Episodic is often assumed by policy gradient methods and also assumed by some model-based RL methods. Infinite horizon means there is no clear endpoint for a trajectory. The time step for the trajectory can go to infinity, but there will be some point where the distribution of the state action pairs remain stable and not changing anymore. We refer this as stationary distribution. Most of the applications we discuss in this paper are episodic and have clear end-points(i.e., mortality, diagnosis of disease). The episodic assumption is usually assumed by pure policy gradient RL and some model-based RL algorithms. Whereas we observed value-based RL algorithm also works decently with many clinical applications in this paper.\\

\textbf{Fully observed/ partially observed MDP}  When the MDP is fully observed, all the main RL algorithms can be applied. Whereas for partially observed MDP, one possible solution is to use recurrent RL, such as LSTM based DRQN algorithm, to aggregate all historical observation as the \textit{belief} state. In real time, most of the clinical applications are POMDP; we are only able to represent a patient's state by their physiological features. For the methods working in POMDPs, the approach of maintaining a \textit{belief state} is widely used besides RNN. The \textit{belief state} is the posterior distribution over latent state based on the historical incomplete and noisy observations. \cite{mcallister2017data} illustrated a particular case where partial observability is due to additive Gaussian white noise on the unobserved state. Then \textit{belief} can be used to filter the noisy observations. \cite{igl2018deep} proposed a deep variational reinforcement learning (DVRL) algorithm that used a DNN to directly output \textit{belief state} from noisy observations. They showed their algorithm outperformed recurrent RL on inferring the actual \textit{belief state}.\\

\section{Challenges and Remedies}

\textbf{Learning from limited observational data}\\
The applications of Deep RL in the clinical setting is very different from the case for the Atari game, where one can repeat the game many many times and play out all possible scenarios to allow the Deep RL agent to learn the optimal policy. In the clinical setting, the Deep RL agent requires to learn from a limited set of data and intervention variations that were collected.
It is known as the POMDP problem. 
Therefore, for clinical applications, the improved policy learned by the RL agent is often not the optimal policy. 
As discussed in the last section, this problem can be addressed by using a recurrent structure such as LSTM\citep{futoma2018learning} or by inferring and maintaining a \textit{belief state} with DNN\citep{igl2018deep}.
\\

\textbf{Definition of state action, reward space for clinical applications}\\
Finding appropriate representation of the state, action, and reward function is challenging in the clinical context\citep{futoma2018learning}. One needs to define the reward function to balance the trade-offs between short-term improvement and long-term success. Take sepsis patients in ICU as an example. Periodic improvements in blood pressure may not lead to improvements in patients' outcome. However, if we solely focus on patients' outcome (survival or death) as the reward, this will result in a very long sequence of learning without any feedbacks for the agent.
%
%Therefore, it is hard to design an appropriate reward function. 
%
While the good news is for some RL algorithms, such as Inverse RL, we do not need to design the reward by hand. It can be approximated using DNN, and we might even train a reward that is closer to the actual reward than hand-crafted reward.\\

\textbf{Performance Benchmarking}\\
%This is a challenge when bridging the gap between research and clinical practice. Because the ground-truth is unknown, researchers struggle to evaluate model predictions in the absence of true labels for optimal decision makings. 
%
In other domain, such as video games,  the successful implementation of DQN on Atari game has attracted the great interest of researches in this field \cite{mnih2013playing}. 
%Many global companies tried using RL methods for video game playing. 
%
For instance, DeepMind applied actor-critic RL for their video game 'StarCraft II'\citep{alghanem2018asynchronous}. Microsoft developed an open source environment for researchers to test on their video game 'Malmo'\citep{johnson2016malmo}.
All these successful implementations now are serving as the benchmarks for any new applications of RL in video games.
However, in the healthcare domain, the benchmark is absent due to the lack of many successful applications. We observed that the majority of the existing RL healthcare applications utilized the MIMIC EHR database \cite{johnson2016mimic}. Thus, we plan to build a set benchmark using the MIMIC data for the application of Deep RL for clinical decision support in ICUs.\\

\textbf{Exploration/ Exploitation} \\
The fundamental dilemma for RL is exploration versus exploitation. If an agent knows to take certain actions would result in good reward, how can the agent decide whether to attempt new behaviors to discover ones with higher reward (Exploration) or continue to do the best thing it knows so far (Exploitation). Exploitation means to do things we know will yield the highest reward, whereas exploration means to do things we have not done before, but in the hopes of getting an even higher reward. 
Exploration can be challenging in clinical settings due to ethics and treatment safety considerations.
One paradigm for exploration-exploitation balancing is to use $\epsilon$-greedy search to explore random action with a probability of $\epsilon \in [0,1]$. The higher the value of $\epsilon$, the more 'open-minded' an agent will be to explore an arbitrary action. Alternatives for choosing exploration or exploitation include optimistic exploration\citep{auer2002finite}, Thompson sampling\citep{chapelle2011empirical} and information gain\citep{mohamed2015variational,houthooft2016vime}.\\

\textbf{Data deficiency and Data Quality}\\
Almost all deep learning models have the problem of data deficiency in healthcare applications. Though there exist available public database, the smaller medical institution often lack sufficient data to fit a good deep learning model relying on their local database. Possible solutions include using GAN based models to generate data from similar distribution\citep{tseng2017deep}, or use transfer learning\citep{haarnoja2017reinforcement} to pre-train the DNN model from larger datasets and later apply it to a smaller hospital/institutional clinical data.

\begin{table}[htbp]
	\centering
	\caption{Deep Reinforcement Learning Applications for Clinical Decision Support}
	
	\begin{tabular}{|p{3.5cm}|p{3.5cm}|l|p{3cm}| p{1.8cm}|}
		
		\hline
		\textbf{Application Context} & \textbf{Task/Treatment} & \textbf{Method}  & \textbf{Data}  & \textbf{Reference}\\ \hline
		Weaning of Mechanical Ventilation  & On/Off Mechanical Ventilation, Sedation dosage & Fitted Q Iteration & MIMIC III & \citep{prasad2017reinforcement}\\ \hline
		Optimal Heparin Dosing in ICU & Medication Dosage  & Fitted Q Iteration & MIMIC II & \citep{nemati2016optimal} \\ 
		\hline
		Concept extraction from Free Clinical Text  & Extract diagnoses and clinical concept& DQN & TREC CDS  & \citep{ling2017diagnostic}\\ 
		\hline
		Symptom Checking 1.0  &imitate behavior of inquiry/diagnosis  &  DQN & Simulated Data & \citep{tang2016inquire} \\ 
		\hline
		Symptom Checking 2.0   & imitate behavior of inquiry/diagnosis  & Hierarchical RL & SymCAT (Simulated) & \citep{kao2018context} \\ 
		\hline
		Sepsis Treatment 1.0 &Vasopressors, IV fluid  & Double DQN & MIMIC III & \citep{raghu2017continuous} \\ 
		\hline
		Sepsis Treatment 2.0  &Vasopressors, IV fluid , antibiotics& MPG+ DRQN & Duke University health system (private)& \citep{futoma2018learning} \\ 
		\hline
		Optimal medical prescription in ICU  &Medication choice & Actor-critic RLL &MIMIC III& \citep{wang2018supervised} \\
		\hline
		Radiation dose fraction for lung cancer&  Chemotherapy, Medication type and dosage & Model-based RL & CIBMTR registry & \citep{tseng2017deep} \\ 
		\hline
	
		Diabetes Treatment   &Investigate reward for patient's state & Inverse RL (Non-DNN) & University of Tokuo Hospital (private)& \citep{asoh2013application} \\ 
		\hline
		
	\end{tabular}
	\label{tab:example} 
\end{table}

\clearpage
\bibliographystyle{unsrt} 
\bibliography{sample_29_Mar}

\begin{thebibliography}{64}
\providecommand{\natexlab}[1]{#1}
\providecommand{\url}[1]{\texttt{#1}}
\expandafter\ifx\csname urlstyle\endcsname\relax
  \providecommand{\doi}[1]{doi: #1}\else
  \providecommand{\doi}{doi: \begingroup \urlstyle{rm}\Url}\fi

\bibitem[Abbeel and Ng(2004)]{abbeel2004apprenticeship}
Pieter Abbeel and Andrew~Y Ng.
\newblock Apprenticeship learning via inverse reinforcement learning.
\newblock In \emph{Proceedings of the twenty-first international conference on
  Machine learning}, page~1. ACM, 2004.

\bibitem[Alghanem et~al.(2018)]{alghanem2018asynchronous}
Basel Alghanem et~al.
\newblock Asynchronous advantage actor-critic agent for starcraft ii.
\newblock \emph{arXiv preprint arXiv:1807.08217}, 2018.

\bibitem[Aronson(2006)]{aronson2006metamap}
Alan~R Aronson.
\newblock Metamap: Mapping text to the umls metathesaurus.
\newblock \emph{Bethesda, MD: NLM, NIH, DHHS}, 1:\penalty0 26, 2006.

\bibitem[Arulkumaran et~al.(2017)Arulkumaran, Deisenroth, Brundage, and
  Bharath]{arulkumaran2017brief}
Kai Arulkumaran, Marc~Peter Deisenroth, Miles Brundage, and Anil~Anthony
  Bharath.
\newblock A brief survey of deep reinforcement learning.
\newblock \emph{arXiv preprint arXiv:1708.05866}, 2017.

\bibitem[Asoh et~al.(2013)Asoh, Akaho, Kamishima, Hasida, Aramaki, and
  Kohro]{asoh2013application}
Hideki Asoh, Masanori Shiro1~Shotaro Akaho, Toshihiro Kamishima, Koiti Hasida,
  Eiji Aramaki, and Takahide Kohro.
\newblock An application of inverse reinforcement learning to medical records
  of diabetes treatment.
\newblock In \emph{ECMLPKDD2013 workshop on reinforcement learning with
  generalized feedback}, 2013.

\bibitem[Auer et~al.(2002)Auer, Cesa-Bianchi, and Fischer]{auer2002finite}
Peter Auer, Nicolo Cesa-Bianchi, and Paul Fischer.
\newblock Finite-time analysis of the multiarmed bandit problem.
\newblock \emph{Machine learning}, 47\penalty0 (2-3):\penalty0 235--256, 2002.

\bibitem[Baird~III(1993)]{baird1993advantage}
Leemon~C Baird~III.
\newblock Advantage updating.
\newblock Technical report, WRIGHT LAB WRIGHT-PATTERSON AFB OH, 1993.

\bibitem[Bottou(2010)]{bottou2010large}
L{\'e}on Bottou.
\newblock Large-scale machine learning with stochastic gradient descent.
\newblock In \emph{Proceedings of COMPSTAT'2010}, pages 177--186. Springer,
  2010.

\bibitem[Chapelle and Li(2011)]{chapelle2011empirical}
Olivier Chapelle and Lihong Li.
\newblock An empirical evaluation of thompson sampling.
\newblock In \emph{Advances in neural information processing systems}, pages
  2249--2257, 2011.

\bibitem[Chernova and Veloso(2007)]{chernova2007confidence}
Sonia Chernova and Manuela Veloso.
\newblock Confidence-based policy learning from demonstration using gaussian
  mixture models.
\newblock In \emph{Proceedings of the 6th international joint conference on
  Autonomous agents and multiagent systems}, page 233. ACM, 2007.

\bibitem[Cire{\c{s}}an et~al.(2012)Cire{\c{s}}an, Meier, and
  Schmidhuber]{cirecsan2012multi}
Dan Cire{\c{s}}an, Ueli Meier, and J{\"u}rgen Schmidhuber.
\newblock Multi-column deep neural networks for image classification.
\newblock \emph{arXiv preprint arXiv:1202.2745}, 2012.

\bibitem[Deisenroth and Rasmussen(2011)]{deisenroth2011pilco}
Marc Deisenroth and Carl~E Rasmussen.
\newblock Pilco: A model-based and data-efficient approach to policy search.
\newblock In \emph{Proceedings of the 28th International Conference on machine
  learning (ICML-11)}, pages 465--472, 2011.

\bibitem[Doya et~al.(2002)Doya, Samejima, Katagiri, and
  Kawato]{doya2002multiple}
Kenji Doya, Kazuyuki Samejima, Ken-ichi Katagiri, and Mitsuo Kawato.
\newblock Multiple model-based reinforcement learning.
\newblock \emph{Neural computation}, 14\penalty0 (6):\penalty0 1347--1369,
  2002.

\bibitem[Futoma et~al.(2018)Futoma, Lin, Sendak, Bedoya, Clement, O'Brien, and
  Heller]{futoma2018learning}
Joseph Futoma, Anthony Lin, Mark Sendak, Armando Bedoya, Meredith Clement, Cara
  O'Brien, and Katherine Heller.
\newblock Learning to treat sepsis with multi-output gaussian process deep
  recurrent q-networks.
\newblock 2018.

\bibitem[Ghavamzadeh et~al.(2015)Ghavamzadeh, Mannor, Pineau, Tamar,
  et~al.]{ghavamzadeh2015bayesian}
Mohammad Ghavamzadeh, Shie Mannor, Joelle Pineau, Aviv Tamar, et~al.
\newblock Bayesian reinforcement learning: A survey.
\newblock \emph{Foundations and Trends{\textregistered} in Machine Learning},
  8\penalty0 (5-6):\penalty0 359--483, 2015.

\bibitem[Goodfellow et~al.(2014)Goodfellow, Pouget-Abadie, Mirza, Xu,
  Warde-Farley, Ozair, Courville, and Bengio]{goodfellow2014generative}
Ian Goodfellow, Jean Pouget-Abadie, Mehdi Mirza, Bing Xu, David Warde-Farley,
  Sherjil Ozair, Aaron Courville, and Yoshua Bengio.
\newblock Generative adversarial nets.
\newblock In \emph{Advances in neural information processing systems}, pages
  2672--2680, 2014.

\bibitem[Grondman et~al.(2012)Grondman, Busoniu, Lopes, and
  Babuska]{grondman2012survey}
Ivo Grondman, Lucian Busoniu, Gabriel~AD Lopes, and Robert Babuska.
\newblock A survey of actor-critic reinforcement learning: Standard and natural
  policy gradients.
\newblock \emph{IEEE Transactions on Systems, Man, and Cybernetics, Part C
  (Applications and Reviews)}, 42\penalty0 (6):\penalty0 1291--1307, 2012.

\bibitem[Haarnoja et~al.(2017)Haarnoja, Tang, Abbeel, and
  Levine]{haarnoja2017reinforcement}
Tuomas Haarnoja, Haoran Tang, Pieter Abbeel, and Sergey Levine.
\newblock Reinforcement learning with deep energy-based policies.
\newblock In \emph{Proceedings of the 34th International Conference on Machine
  Learning-Volume 70}, pages 1352--1361. JMLR. org, 2017.

\bibitem[Harmon and Baird~III(1996)]{harmon1996multi}
Mance~E Harmon and Leemon~C Baird~III.
\newblock Multi-player residual advantage learning with general function
  approximation.
\newblock \emph{Wright Laboratory, WL/AACF, Wright-Patterson Air Force Base,
  OH}, pages 45433--7308, 1996.

\bibitem[Hasselt(2010)]{hasselt2010double}
Hado~V Hasselt.
\newblock Double q-learning.
\newblock In \emph{Advances in Neural Information Processing Systems}, pages
  2613--2621, 2010.

\bibitem[Hausknecht and Stone(2015)]{hausknecht2015deep}
Matthew Hausknecht and Peter Stone.
\newblock Deep recurrent q-learning for partially observable mdps.
\newblock In \emph{2015 AAAI Fall Symposium Series}, 2015.

\bibitem[Heess et~al.(2015)Heess, Wayne, Silver, Lillicrap, Erez, and
  Tassa]{heess2015learning}
Nicolas Heess, Gregory Wayne, David Silver, Timothy Lillicrap, Tom Erez, and
  Yuval Tassa.
\newblock Learning continuous control policies by stochastic value gradients.
\newblock In \emph{Advances in Neural Information Processing Systems}, pages
  2944--2952, 2015.

\bibitem[Hochreiter and Schmidhuber(1997)]{hochreiter1997long}
Sepp Hochreiter and J{\"u}rgen Schmidhuber.
\newblock Long short-term memory.
\newblock \emph{Neural computation}, 9\penalty0 (8):\penalty0 1735--1780, 1997.

\bibitem[Houthooft et~al.(2016)Houthooft, Chen, Duan, Schulman, De~Turck, and
  Abbeel]{houthooft2016vime}
Rein Houthooft, Xi~Chen, Yan Duan, John Schulman, Filip De~Turck, and Pieter
  Abbeel.
\newblock Vime: Variational information maximizing exploration.
\newblock In \emph{Advances in Neural Information Processing Systems}, pages
  1109--1117, 2016.

\bibitem[Howard(1960)]{howard1960dynamic}
Ronald~A Howard.
\newblock Dynamic programming and markov processes.
\newblock 1960.

\bibitem[Igl et~al.(2018)Igl, Zintgraf, Le, Wood, and Whiteson]{igl2018deep}
Maximilian Igl, Luisa Zintgraf, Tuan~Anh Le, Frank Wood, and Shimon Whiteson.
\newblock Deep variational reinforcement learning for pomdps.
\newblock \emph{arXiv preprint arXiv:1806.02426}, 2018.

\bibitem[Johnson et~al.(2016{\natexlab{a}})Johnson, Pollard, Shen, Li-wei,
  Feng, Ghassemi, Moody, Szolovits, Celi, and Mark]{johnson2016mimic}
Alistair~EW Johnson, Tom~J Pollard, Lu~Shen, H~Lehman Li-wei, Mengling Feng,
  Mohammad Ghassemi, Benjamin Moody, Peter Szolovits, Leo~Anthony Celi, and
  Roger~G Mark.
\newblock Mimic-iii, a freely accessible critical care database.
\newblock \emph{Scientific data}, 3:\penalty0 160035, 2016{\natexlab{a}}.

\bibitem[Johnson et~al.(2016{\natexlab{b}})Johnson, Hofmann, Hutton, and
  Bignell]{johnson2016malmo}
Matthew Johnson, Katja Hofmann, Tim Hutton, and David Bignell.
\newblock The malmo platform for artificial intelligence experimentation.
\newblock In \emph{IJCAI}, pages 4246--4247, 2016{\natexlab{b}}.

\bibitem[Kaelbling et~al.(1996)Kaelbling, Littman, and
  Moore]{kaelbling1996reinforcement}
Leslie~Pack Kaelbling, Michael~L Littman, and Andrew~W Moore.
\newblock Reinforcement learning: A survey.
\newblock \emph{Journal of artificial intelligence research}, 4:\penalty0
  237--285, 1996.

\bibitem[Kaelbling et~al.(1998)Kaelbling, Littman, and
  Cassandra]{kaelbling1998planning}
Leslie~Pack Kaelbling, Michael~L Littman, and Anthony~R Cassandra.
\newblock Planning and acting in partially observable stochastic domains.
\newblock \emph{Artificial intelligence}, 101\penalty0 (1-2):\penalty0 99--134,
  1998.

\bibitem[Kao et~al.(2018)Kao, Tang, and Chang]{kao2018context}
Hao-Cheng Kao, Kai-Fu Tang, and Edward~Y Chang.
\newblock Context-aware symptom checking for disease diagnosis using
  hierarchical reinforcement learning.
\newblock In \emph{Thirty-Second AAAI Conference on Artificial Intelligence},
  2018.

\bibitem[Kapadia(1998)]{kapadia1998discriminative}
Sadik Kapadia.
\newblock \emph{Discriminative training of hidden Markov models}.
\newblock PhD thesis, Citeseer, 1998.

\bibitem[Kiumarsi et~al.(2018)Kiumarsi, Vamvoudakis, Modares, and
  Lewis]{kiumarsi2018optimal}
Bahare Kiumarsi, Kyriakos~G Vamvoudakis, Hamidreza Modares, and Frank~L Lewis.
\newblock Optimal and autonomous control using reinforcement learning: A
  survey.
\newblock \emph{IEEE transactions on neural networks and learning systems},
  29\penalty0 (6):\penalty0 2042--2062, 2018.

\bibitem[Krstic et~al.(1995)Krstic, Kanellakopoulos, Kokotovic,
  et~al.]{krstic1995nonlinear}
Miroslav Krstic, Ioannis Kanellakopoulos, Petar~V Kokotovic, et~al.
\newblock \emph{Nonlinear and adaptive control design}, volume 222.
\newblock Wiley New York, 1995.

\bibitem[Kulkarni et~al.(2016)Kulkarni, Narasimhan, Saeedi, and
  Tenenbaum]{kulkarni2016hierarchical}
Tejas~D Kulkarni, Karthik Narasimhan, Ardavan Saeedi, and Josh Tenenbaum.
\newblock Hierarchical deep reinforcement learning: Integrating temporal
  abstraction and intrinsic motivation.
\newblock In \emph{Advances in neural information processing systems}, pages
  3675--3683, 2016.

\bibitem[Lin(1992)]{lin1992self}
Long-Ji Lin.
\newblock Self-improving reactive agents based on reinforcement learning,
  planning and teaching.
\newblock \emph{Machine learning}, 8\penalty0 (3-4):\penalty0 293--321, 1992.

\bibitem[Ling et~al.(2017)Ling, Hasan, Datla, Qadir, Lee, Liu, and
  Farri]{ling2017diagnostic}
Yuan Ling, Sadid~A Hasan, Vivek Datla, Ashequl Qadir, Kathy Lee, Joey Liu, and
  Oladimeji Farri.
\newblock Diagnostic inferencing via improving clinical concept extraction with
  deep reinforcement learning: A preliminary study.
\newblock In \emph{Machine Learning for Healthcare Conference}, pages 271--285,
  2017.

\bibitem[Marik(2015)]{marik2015demise}
PE~Marik.
\newblock The demise of early goal-directed therapy for severe sepsis and
  septic shock.
\newblock \emph{Acta Anaesthesiologica Scandinavica}, 59\penalty0 (5):\penalty0
  561--567, 2015.

\bibitem[McAllister and Rasmussen(2017)]{mcallister2017data}
Rowan McAllister and Carl~Edward Rasmussen.
\newblock Data-efficient reinforcement learning in continuous state-action
  gaussian-pomdps.
\newblock In \emph{Advances in Neural Information Processing Systems}, pages
  2040--2049, 2017.

\bibitem[Mnih et~al.(2013)Mnih, Kavukcuoglu, Silver, Graves, Antonoglou,
  Wierstra, and Riedmiller]{mnih2013playing}
Volodymyr Mnih, Koray Kavukcuoglu, David Silver, Alex Graves, Ioannis
  Antonoglou, Daan Wierstra, and Martin Riedmiller.
\newblock Playing atari with deep reinforcement learning.
\newblock \emph{arXiv preprint arXiv:1312.5602}, 2013.

\bibitem[Mnih et~al.(2015)Mnih, Kavukcuoglu, Silver, Rusu, Veness, Bellemare,
  Graves, Riedmiller, Fidjeland, Ostrovski, et~al.]{mnih2015human}
Volodymyr Mnih, Koray Kavukcuoglu, David Silver, Andrei~A Rusu, Joel Veness,
  Marc~G Bellemare, Alex Graves, Martin Riedmiller, Andreas~K Fidjeland, Georg
  Ostrovski, et~al.
\newblock Human-level control through deep reinforcement learning.
\newblock \emph{Nature}, 518\penalty0 (7540):\penalty0 529, 2015.

\bibitem[Mnih et~al.(2016)Mnih, Badia, Mirza, Graves, Lillicrap, Harley,
  Silver, and Kavukcuoglu]{mnih2016asynchronous}
Volodymyr Mnih, Adria~Puigdomenech Badia, Mehdi Mirza, Alex Graves, Timothy
  Lillicrap, Tim Harley, David Silver, and Koray Kavukcuoglu.
\newblock Asynchronous methods for deep reinforcement learning.
\newblock In \emph{International conference on machine learning}, pages
  1928--1937, 2016.

\bibitem[Mohamed and Rezende(2015)]{mohamed2015variational}
Shakir Mohamed and Danilo~Jimenez Rezende.
\newblock Variational information maximisation for intrinsically motivated
  reinforcement learning.
\newblock In \emph{Advances in neural information processing systems}, pages
  2125--2133, 2015.

\bibitem[Munos and Szepesv{\'a}ri(2008)]{munos2008finite}
R{\'e}mi Munos and Csaba Szepesv{\'a}ri.
\newblock Finite-time bounds for fitted value iteration.
\newblock \emph{Journal of Machine Learning Research}, 9\penalty0
  (May):\penalty0 815--857, 2008.

\bibitem[Nemati et~al.(2016)Nemati, Ghassemi, and Clifford]{nemati2016optimal}
Shamim Nemati, Mohammad~M Ghassemi, and Gari~D Clifford.
\newblock Optimal medication dosing from suboptimal clinical examples: A deep
  reinforcement learning approach.
\newblock In \emph{2016 38th Annual International Conference of the IEEE
  Engineering in Medicine and Biology Society (EMBC)}, pages 2978--2981. IEEE,
  2016.

\bibitem[Ng et~al.(2000)Ng, Russell, et~al.]{ng2000algorithms}
Andrew~Y Ng, Stuart~J Russell, et~al.
\newblock Algorithms for inverse reinforcement learning.
\newblock In \emph{Icml}, volume~1, page~2, 2000.

\bibitem[Pan and Yang(2010)]{pan2010survey}
Sinno~Jialin Pan and Qiang Yang.
\newblock A survey on transfer learning.
\newblock \emph{IEEE Transactions on knowledge and data engineering},
  22\penalty0 (10):\penalty0 1345--1359, 2010.

\bibitem[Prasad et~al.(2017)Prasad, Cheng, Chivers, Draugelis, and
  Engelhardt]{prasad2017reinforcement}
Niranjani Prasad, Li-Fang Cheng, Corey Chivers, Michael Draugelis, and
  Barbara~E Engelhardt.
\newblock A reinforcement learning approach to weaning of mechanical
  ventilation in intensive care units.
\newblock \emph{arXiv preprint arXiv:1704.06300}, 2017.

\bibitem[Raghu et~al.(2017)Raghu, Komorowski, Celi, Szolovits, and
  Ghassemi]{raghu2017continuous}
Aniruddh Raghu, Matthieu Komorowski, Leo~Anthony Celi, Peter Szolovits, and
  Marzyeh Ghassemi.
\newblock Continuous state-space models for optimal sepsis treatment-a deep
  reinforcement learning approach.
\newblock \emph{arXiv preprint arXiv:1705.08422}, 2017.

\bibitem[Rasmussen(2003)]{rasmussen2003gaussian}
Carl~Edward Rasmussen.
\newblock Gaussian processes in machine learning.
\newblock In \emph{Summer School on Machine Learning}, pages 63--71. Springer,
  2003.

\bibitem[Riedmiller(2005)]{riedmiller2005neural}
Martin Riedmiller.
\newblock Neural fitted q iteration--first experiences with a data efficient
  neural reinforcement learning method.
\newblock In \emph{European Conference on Machine Learning}, pages 317--328.
  Springer, 2005.

\bibitem[Schaal(1999)]{schaal1999imitation}
Stefan Schaal.
\newblock Is imitation learning the route to humanoid robots?
\newblock \emph{Trends in cognitive sciences}, 3\penalty0 (6):\penalty0
  233--242, 1999.

\bibitem[Silver et~al.(2016)Silver, Huang, Maddison, Guez, Sifre, Van
  Den~Driessche, Schrittwieser, Antonoglou, Panneershelvam, Lanctot,
  et~al.]{silver2016mastering}
David Silver, Aja Huang, Chris~J Maddison, Arthur Guez, Laurent Sifre, George
  Van Den~Driessche, Julian Schrittwieser, Ioannis Antonoglou, Veda
  Panneershelvam, Marc Lanctot, et~al.
\newblock Mastering the game of go with deep neural networks and tree search.
\newblock \emph{nature}, 529\penalty0 (7587):\penalty0 484, 2016.

\bibitem[Simpson et~al.(2014)Simpson, Voorhees, and Hersh]{simpson2014overview}
Matthew~S Simpson, Ellen~M Voorhees, and William Hersh.
\newblock Overview of the trec 2014 clinical decision support track.
\newblock Technical report, LISTER HILL NATIONAL CENTER FOR BIOMEDICAL
  COMMUNICATIONS BETHESDA MD, 2014.

\bibitem[Sutton and Barto(2018)]{sutton2018reinforcement}
Richard~S Sutton and Andrew~G Barto.
\newblock \emph{Reinforcement learning: An introduction}.
\newblock MIT press, 2018.

\bibitem[Sutton et~al.(1998)Sutton, Barto, et~al.]{sutton1998introduction}
Richard~S Sutton, Andrew~G Barto, et~al.
\newblock \emph{Introduction to reinforcement learning}, volume 135.
\newblock MIT press Cambridge, 1998.

\bibitem[Sutton et~al.(2000)Sutton, McAllester, Singh, and
  Mansour]{sutton2000policy}
Richard~S Sutton, David~A McAllester, Satinder~P Singh, and Yishay Mansour.
\newblock Policy gradient methods for reinforcement learning with function
  approximation.
\newblock In \emph{Advances in neural information processing systems}, pages
  1057--1063, 2000.

\bibitem[Tang et~al.(2016)Tang, Kao, Chou, and Chang]{tang2016inquire}
Kai-Fu Tang, Hao-Cheng Kao, Chun-Nan Chou, and Edward~Y Chang.
\newblock Inquire and diagnose: Neural symptom checking ensemble using deep
  reinforcement learning.
\newblock In \emph{Proceedings of NIPS Workshop on Deep Reinforcement
  Learning}, 2016.

\bibitem[Thrun and Schwartz(1993)]{thrun1993issues}
Sebastian Thrun and Anton Schwartz.
\newblock Issues in using function approximation for reinforcement learning.
\newblock In \emph{Proceedings of the 1993 Connectionist Models Summer School
  Hillsdale, NJ. Lawrence Erlbaum}, 1993.

\bibitem[Tseng et~al.(2017)Tseng, Luo, Cui, Chien, Ten~Haken, and
  Naqa]{tseng2017deep}
Huan-Hsin Tseng, Yi~Luo, Sunan Cui, Jen-Tzung Chien, Randall~K Ten~Haken, and
  Issam~El Naqa.
\newblock Deep reinforcement learning for automated radiation adaptation in
  lung cancer.
\newblock \emph{Medical physics}, 44\penalty0 (12):\penalty0 6690--6705, 2017.

\bibitem[van Hasselt(2011)]{van2011insights}
Hado~Philip van Hasselt.
\newblock \emph{Insights in reinforcement learning}.
\newblock Hado van Hasselt, 2011.

\bibitem[Wang et~al.(2018)Wang, Zhang, He, and Zha]{wang2018supervised}
Lu~Wang, Wei Zhang, Xiaofeng He, and Hongyuan Zha.
\newblock Supervised reinforcement learning with recurrent neural network for
  dynamic treatment recommendation.
\newblock In \emph{Proceedings of the 24th ACM SIGKDD International Conference
  on Knowledge Discovery \& Data Mining}, pages 2447--2456. ACM, 2018.

\bibitem[Wang et~al.(2015)Wang, Schaul, Hessel, Van~Hasselt, Lanctot, and
  De~Freitas]{wang2015dueling}
Ziyu Wang, Tom Schaul, Matteo Hessel, Hado Van~Hasselt, Marc Lanctot, and Nando
  De~Freitas.
\newblock Dueling network architectures for deep reinforcement learning.
\newblock \emph{arXiv preprint arXiv:1511.06581}, 2015.

\bibitem[Watkins and Dayan(1992)]{watkins1992q}
Christopher~JCH Watkins and Peter Dayan.
\newblock Q-learning.
\newblock \emph{Machine learning}, 8\penalty0 (3-4):\penalty0 279--292, 1992.

\end{thebibliography}

\end{document}